# Residual-SwinCA-Net: A Channel-Aware Integrated Residual CNN–Swin Transformer for Malignant Lesion Segmentation in BUSI

Saeeda Naz[1], Saddam Hussain Khan[1]*

[1]Artificial Intelligence Lab, Department of Computer Systems Engineering, University of Engineering and Applied Sciences (UEAS), Swat 19060, Pakistan

**Email:** hengrshkhan822@gmail.com

Abstract

The main reason behind the early and precise detection of breast cancer in women is the leading cause worldwide. Manual detection systems of such kind of cancer provide inconsistency due to heterogeneous lesion morphology, speckle interference, low contrast, indistinct boundaries, and noise in ultrasound images, and are a time-consuming procedure. Therefore, a novel deep hybrid 'Residual-SwinCA-Net' segmentation framework is proposed in the study for addressing such kind of challenges by extracting locally correlated and robust features incorporated by residual CNN modules. Furthermore, for learning global dependencies, Swin Transformer blocks are customized using internal residual pathways, reinforcing gradient stability, refining local patterns, and global feature fusion. Formerly, for enhancing tissue continuity, ultrasound noise suppressions, and accentuating fine structural transitions Laplacian-of-Gaussian regional operator is applied, and for maintaining the morphological integrity of malignant lesion contours, a boundary-oriented operator has been incorporated. Subsequently, a contraction strategy was applied stage-wise by progressively reducing features-map progressively for capturing scale invariance and enhancing the robustness of structural variability. In addition, each decoder level prior augmentation integrates a new Multi-Scale Channel Attention and Squeezing (MSCAS) module. The MSCAS selectively emphasizes encoder salient maps, retains discriminative global context, and complementary local structures with minimal computational cost while suppressing redundant activations. Finally, the Pixel-Attention module encodes class-relevant spatial cues by adaptively weighing malignant lesion pixels while suppressing background interference. The Residual-SwinCA-Net and existing CNNs/ViTs techniques have been implemented on the publicly available BUSI dataset. The proposed Residual-SwinCA-Net framework outperformed and achieved 99.29% mean accuracy, 98.74% IoU, and 0.9041 Dice for breast lesion segmentation. The proposed Residual-SwinCA-Net framework improves the BUSI lesion diagnostic performance and strengthens timely clinical decision-making.

Keywords: Breast Cancer, Ultrasound, CNN, Transformer, Swin Transformer, Attention

# 1 Introduction

Breast cancer is the second leading cause of death globally, and the most affordable, safe, and effective way to analyze malignant tumors is the ultrasound diagnostic tools [1]. Manually examining may lead to the missing lesion region in the breast ultrasound image (BUSI), and may lead to misinterpretation by the radiologists [2]. Emerging and automatic technologies will enhance diagnostic consistency and improve patient care by detecting the lesion region reliably and accurately. Region growing [3], threshold [4], and clustering [5] traditional techniques will lead to inaccurate results due to their handcrafted feature extractions and noise-sensitive artifacts with BUSI.

Extracting meaningful features directly from the data using deep learning (DL) offers an effective solution by learning hierarchical features. Encoder-decoder concepts with skip connections used in UNet [6] and its variants like UNet++ [7], U2Net [8], and SA-UNet [9], have enhanced segmentation accuracy and localization in image-based segmentation problems. CNNs learn correlated features but struggle with global dependencies in complex data [10], therefore, motivated efforts toward expanding receptive fields for stronger regional discrimination [11][12]. In contrast, Transformer backbones like ViT address this limitation through self-attention and driven global context modeling. However, their optimal performance depends on access to large, well-curated training datasets [13].

Addressing these challenges, hybrid models are explored in recent studies integrating both CNNs with the transformer blocks for segmenting medical images, such as TransUNet [14] and LevitUNet [15]. These hybrid networks have solved the issues; however, they still suffer from challenges like CNNs extracting local spatial features at multiple scales, whereas Transformers encode global contexts as tokens, leading to heterogeneity. In medical imaging, limited sample sizes relative to feature dimensionality further exacerbate overfitting. Additionally, BUS datasets are constrained by scarcity, speckle noise, low contrast, and morphological lesions variability. Moreover, the bias of the model towards the non-lesion region, as the lesion region is much smaller than the background. Therefore, domain shifts due to the patient's anatomy, acquisition devices, and imaging modalities will degrade the model's generalization.

In this study, we introduced a novel breast lesion detection technique based on a robust encoder–decoder architecture with integrated multi-stage feature fusion. The proposed 'Residual-SwinCA-Net' segmentation method includes four hierarchical encoders and decoders that progressively identify increasingly complex malignant features. Residual-SwinCA-Net combines early-stage residual CNNs for efficient local feature extraction with a hybrid design to enhance inputs for transformer processing. The key contributions of the research are as follows:

- A novel hybrid 'Residual-SwinCA-Net' segmentation architecture is introduced that systematically integrates residual CNN with customized SwinT blocks to learn fine-grained local lesion cues and long-range contextual dependencies in BUSI. Moreover, skip

- connections are incorporated at each decoder stage to concatenate encoder-decoder local–global feature coupling, enabling robust representation learning in heterogeneous and low-contrast BUSI.
- Laplacian-of-Gaussian regional operator and boundary-aware refinement mechanism are integrated to improve spatial homogeneity, suppress speckle noise, and preserve malignant lesion morphology, particularly along irregular and indistinct boundaries. Additionally, stage-wise down-sampling after each encoder progressively reduces channel resolution to capture scale invariance and strengthen robustness to heterogeneous lesion morphology.
- In addition, each decoder level prior augmentation integrates a new Multi-Scale Channel Attention and Squeezing (MSCAS) block. The MSCAS emphasizes encoder salient maps, retains discriminative global context and local correlated features by reducing computational cost while suppressing redundant activations. Finally, the Pixel-Attention block in the final decoding portion learns class-relevant spatial cues by adaptively weighting malignant lesion pixels while suppressing background interference.
- The proposed Residual-SwinCA-Net framework has been assessed on a standard BUSI dataset and demonstrates outperformance over the existing CNNS/ViTs techniques.

The paper is structured with the previous studies and methodology in sections 2 and 3, respectively. The dataset and experimental setup are presented in Section 4. The results are discussed in Section 5, while the conclusion is illustrated in Section 6.

## 2    Previous Studies

BUS tumor segmentation strategies are divided into conventional methods and DL-based approaches. Conventional methods, such as region-based, deformable models, graph-based, and learning-based techniques, rely on manually engineered features and texture analysis. These foundational techniques have significantly contributed to advancements in modern BUS tumor segmentation. [16] relying on individually made features and texture analysis often fails in complex cases [17]. However, these methods often face challenges in accurately identifying lesions in complex ultrasound images.

### 2.1    CNN-based Breast Lesion Segmentation

Several CNN-based techniques for breast lesion analysis are presented in the literature. However, while [18] introduces a fully convolutional network (FCN) that performs well among traditional methods, it still struggles with capturing multi-scale contextual information. This limitation is addressed by [11] by employing dilated convolutions for expanding the receptive fields. Furthermore, the most discriminative tumor regions cannot highlight sufficiently the dilated convolutions alone; an attention mechanism by [19] is incorporated to better focus on critical areas. Moreover, attention mechanisms applied miss global structural cues still a applied, global guidance block introduced in [20] for long-range feature representation enhancement, and finally, [21] utilizes a boundary-aware GCN module to further refine edge details.

Multi-scale adaptive feature extraction and contextual correlation have been utilized in [22] to enhance multi-level fusion and reduce the gap between encoder and decoder. Building on the need for stronger feature expression, AAU-Net [23] further improves representation by incorporating hybrid attention modules that refine both spatial and channel emphasis. Similarly, MDF-Net [24] introduces a dual-stage framework with multiscale refinement to improve robustness and segmentation accuracy. Beyond segmentation, lymphocyte identification in immunohistochemistry images has been explored by Rauf et al. through a deep object-detection framework (DC-Lym-AF) [25], while Zafar et al. detect and count tumor-infiltrating lymphocytes using a CNN-based two-phase architecture, TDC-LC [26]. However, despite these advancements, CNN-based models still struggle to learn global-range and contextual details, both of which are essential for accurately segmenting tumors with highly complex shapes and textures.

## 2.2 Transformers-based Breast Lesion Analysis

Transformers have gained significant attention for their strong capability of modeling global dependencies in medical diagnosis [27]. Upon this strength, extracting fine-grained global context, a late fusion strategy combining CNNs for capturing local spatial details with transformers is employed in TransFuse [28]. Further exploring, 1D sequences representation of 3D medical images are passed through stacked transformer encoders represented in UNETR, enabling more comprehensive global feature reasoning [29]. Likewise, leveraging hierarchical window-based attention for both local and global representation learning, a U-shaped architecture specifically tailored for medical imaging tasks by Swin-Unet is extended by Swin Transformer.

For enabling efficient integration of broad global context with fine local details, gated axial attention is combined with the local–global (LoGo) training strategy in the MedT [30]. Building on the need for stronger cross-scale representations, a dual-level feature fusion (DLF) module is incorporated in HiFormer, applying in a hybrid framework for further enhancing feature learning. Additionally, a cross-attention mechanism is employed by HAU-Net in its decoder for strengthening the feature interaction and improving the feature extraction strategy [31].

Both global and local features are integrated in CMFF-Net to improve the performance through a local refinement block and a cross-attention fusion module [32]. Similarly, TransU-Net employs deep supervision and a full-resolution residual stream for enhancing feature learning, while irrelevant features are suppressed by a transformer module for more reliable prediction [33]. A vision transformer combined with a hybrid CNN architecture was used to model long-range dependencies in [34], to further advance global contextual reasoning in medical image segmentation. Moreover, EH-Former incorporates progressive region-wise hard learning, D-Fusion, and uncertainty estimation to better handle challenging tumor regions [35]. Despite these advancements, previous work still exhibits several limitations, including the following:

- Spatial feature extractions by existing CNNs are struggling to capture local spatial regions and have limitations in complex regions.
- Conventional ViT models require great computational costs and extensive training time, and have inefficiency in extracting of local feature extraction, limiting practical deployment.
- Deep networks are facing challenges of generalization, vanishing gradients, and hindering learning efficiency.

The CNN-Transformers have shown effectiveness in medical diagnosis by combining precise local feature extraction with a comprehensive understanding of global context. Implementation of the hybrid-based CNN-Transformer framework, having initial blocks of residual convolution encoding, will effectively extract locally correlated robust features, and Swin-T blocks will extract long-term dependencies in the breast cancer image dataset. Furthermore, MSCAS will attempt to ignore the redundant features, while passing the most relevant informative features will enable more accurate modeling of complex tumor structures.

## 3    Methodology

The proposed DL hybrid 'Residual-SwinCA-Net' breast lesion segmentation technique comprises two residual CNN blocks and two blocks of customized SwinT, which extract local and global context for cancer segmentation. In encoders 1 and 2, residual convolutional blocks extract foundational spatial cues and delineate regional boundaries from the provided images [36]. These two residual stages produce maps at low-level and mid-level, having enriched spatial fidelity and stable flow of gradient, forming a local representation foundation to the model. Consequently, Encoders 3 and 4 incorporate customized SwinT blocks that already operate on refined convolutional space to model global-range dependencies and enhance the framework capacity to resolve subtle malignant patterns within heterogeneous sonographic tissues inside the breast cancer images [37]. This stage will focus on hierarchical global dependencies rather than raw pixel noise. Moreover, the 'Residual-SwinCA-Net' integrates a LoG regional operator after each encoder stage to enhance spatial homogeneity, suppress speckle noise, and preserve heterogeneous malignant lesion morphology [38].

In addition, for improving mask reconstruction and lesion delineation, a novel MSCAS module is incorporated throughout the framework for adaptively reweighting channels, suppressing redundant activations, and retaining diagnostically salient features by assigning weights. At the decoder side, high-resolution encoder features are fused with the up-sampled decoder through skip connections and concatenated to improve final reconstruction. The segmentation mask is reconstructed progressively in a series of 2D up-sampling and convolutional layers for preserving spatial detail, while integrating information regarding multi-scale context. Moreover, class-specific spatial cues are encoded by the pixel attention module to adaptively enhance malignant lesion features while suppressing background regions. Finally, generating an output of a (256 × 256 × 1) segmentation mask with both strong localization abilities from the deep layers and precise boundary delineation for accurate lesion segmentation. The proposed technique and existing ViTs/CNNs are evaluated using the selected dataset and compared using the selected performance metrics. Data augmentation is a preprocessing applied to the input images to mitigate bias and improve generalization. Figure 1 demonstrates the overall methodological setup.

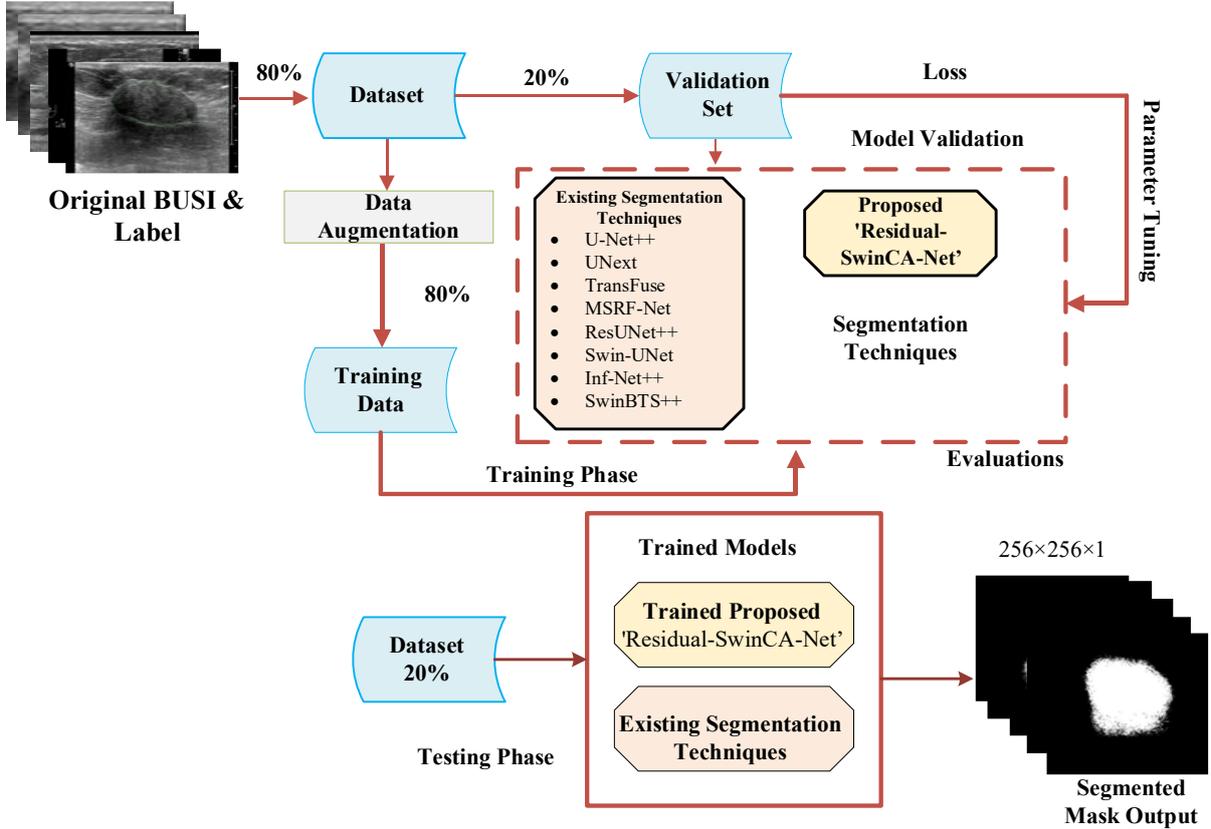

Figure 1: Workflow of the proposed BUSI lesion segmentation.

## 3.1 Multiscale Local and Global Features Extraction

The proposed 'Residual-SwinCA-Net' segmentation framework consists of four stages of encoder: two of them are residual convolution encoder blocks, and the other two stages are SwinT. The first two residual convolution encoder blocks were used to extract initial features with skip connections, mitigating the vanishing gradients problem and improving feature propagation in the network, starting with a (256 × 256 × 3) input [39]. Passing the input image through the residual convolution encoder block 1 for optimal local features having information of malignant tumors patterns and textures, followed by the conv1_relu block, and then average pooling for noise irregularity and morphological features, will reduce the spatial size to (128 × 128) with 64 channels. Furthermore, by passing through stage two, which includes a residual convolution encoder block 2, conv2_relu, and max pooling, the spatial size will be reduced to (64 × 64) with 128 channels.

SwinT [40], [41], is customized in the proposed framework to learn multi-scale feature extraction by utilizing a shifted-window partitioning method for efficient self-attention. The customized SwinT normally takes raw images as input and divides them into small patches for further processing. However, in this study, the SwinT is customized in such a way that it takes optimally local correlated texture features by employing Window Multi-Head Self-Attention (W-MSA), Layer Normalization (LN), and Multi-Layer Perceptron (MLP) for advanced feature representation and modeling long-range dependencies. A feature map of (32 × 32) with 256 channels was obtained by the first transformer block, followed by average-pooling with a window size of 2 in stage 3. And in stages, having a second transformer block followed by a final max-pooling yielding the most abstract feature map at (16 × 16) with 512 channels.

### 3.1.1 Residual Learning for Local Correlated/Texture Variation

A systematic CNN residual approach has been applied to the encoder side of the proposed Residual-SwinCA-Net framework for progressively learning discriminative breast cancer features from image-based datasets, as illustrated in Figure 2. Giving the images directly to the transformer blocks will miss the local information; therefore, we have passed the BUSI through the residual blocks for learning local correlated and texture patterns. This residual block employs a 1×1 convolution for channel-wise transformation, followed by a 3×3 linear projection layer to map the learned representations into richer and more distinct output dimensionalities. For ensuring the adaptive nature preservation of the fine-grained local details, such as subtle texture variations and tumor boundaries, by incorporating a 3×3 kernel size, which is critical for reliable breast cancer analysis. The residual block learn complex BUSI lesion patterns and keeps the gradient flow intact through its skip connection, represented in Equations 1 and 2. These equations describe the link between the input $'x'$ and output '$y$', where the residual term is defined as $T(x, \{w_i\}) = w_2\, \sigma(w_1 x)$ and combined with the input through element-wise addition $T + x$, followed by a ReLU activation, σ. This structure eases optimization in deeper layers and stabilizes training. The input and output sizes differ; that is common with varied lesion variation and multi-scale texture features. A linear projection $w_s$ adjusts the feature dimensions so that spatial and structural details remain consistent across the encoding stages.

$$y = T.(x, \{w_i\}) + x \qquad (1)$$

$$y = T.(x, \{w_i\}) + w_s x \qquad (2)$$

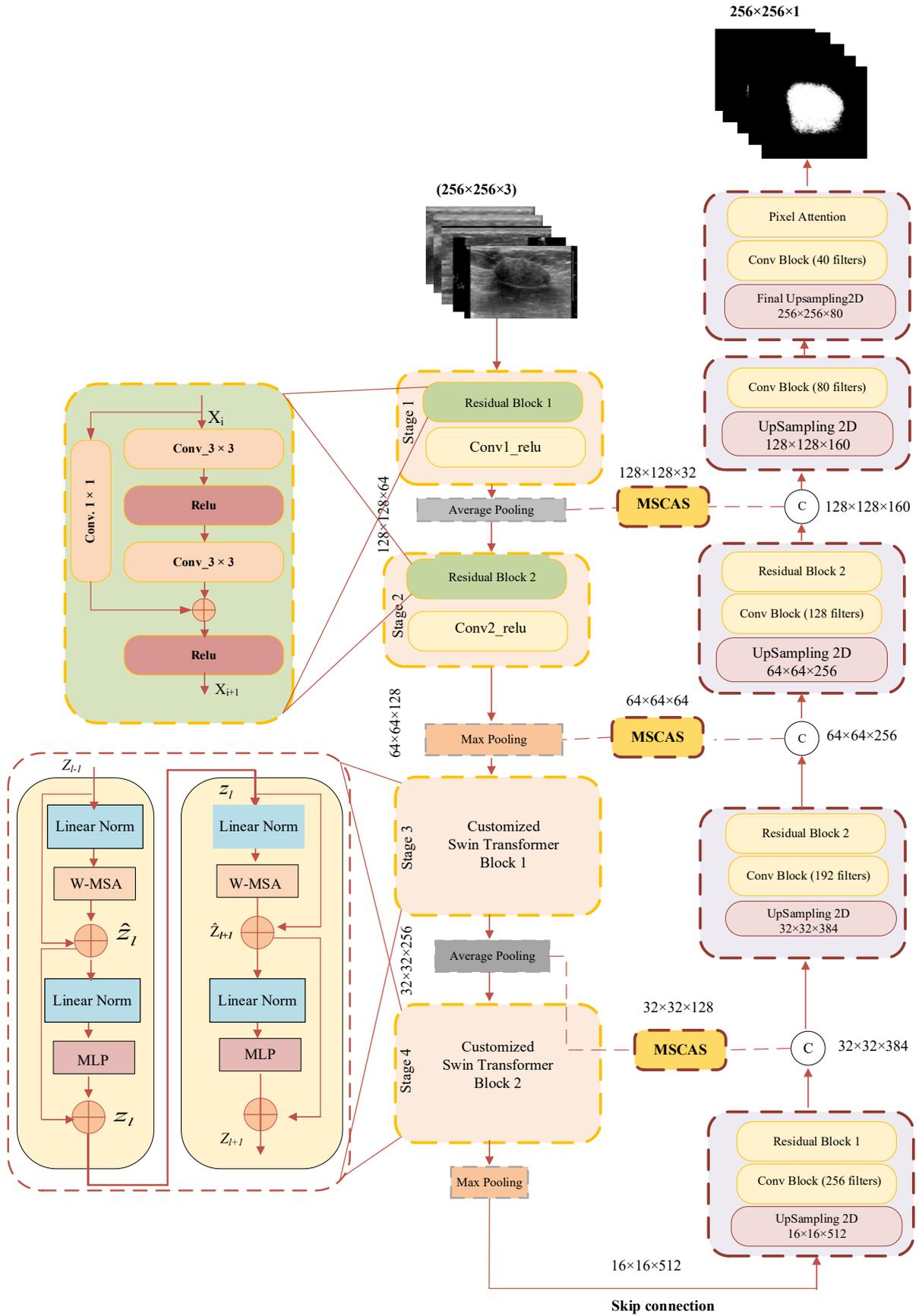

Figure 2: The proposed Residual-SwinCA-Net framework for BUSI lesion segmentation.

### 3.1.2 Customized SwinT for Global Structural Feature Extraction

The transformer encoder consists of a MSA module and a Multi-Layer Perceptron (MLP), both preceded by LN and followed by residual connections, which facilitate stable gradient flow at each stacking layer. The transformer input layers are carefully customized for the imaging domain of the breast cancer dataset in the proposed 'Residual-SwinCA-Net' segmentation framework. Specifically, preceding residual CNN blocks are incorporated to extract locally correlated features. These features comprise a significant BUSI lesion pattern, like dense tissue regions, tumor boundaries, and subtle textural variations. The output $z_l$ of the *l*-th layer thus reflects an integrated understanding of both local details and global context, formally described in Equations 3 and 4. For attending to each token to every other token, providing quadratic computational complexity in the standard Transformer, which makes it inefficient for medical imaging tasks with dense and high-resolution data.

A windowing-W-MSA mechanism incorporated by Swin Transformer to overcome this limitation, which significantly reduces computation by restricting attention to localized regions. The input features map is partitioned into non-overlapping windows of $M \times M$ patches by W-MSA as given in Equation 5. To enable the model to focus more on the localized breast cancer patterns, such as lesion edges, micro-calcification clusters, tissue distortions, and subtle mass textures commonly observed in mammography and ultrasound images, the Self-attention mechanism is then computed independently within each window. This localized attention improves both efficiency and sensitivity to small-scale abnormalities. The outputs produced by the W-MSA module and the subsequent MLP layer are denoted by $\hat{z}_l$ and $z_l$, respectively, with the corresponding notation formally defined in Equation 6.

$$\hat{z}_l = MSA(LN(z_{l-1})) + z_{l-1} \quad (3)$$

$$z_l = MLP(LN(\hat{z}_l)) + \hat{z}_l \quad (4)$$

$$\hat{z}_l = W\_MSA(LN(z_{l-1})) + z_{l-1} \quad (5)$$

$$z_l = MLP(LN(\hat{z}_l)) + \hat{z}_l \quad (6)$$

$$Att.(Q, K, V) = soft.\left(\frac{QK^T}{\sqrt{d}}\right)V \quad (7)$$

Q, K, and V denote the query, key, and value matrices, and represent the dimensionality of the query and key vectors. In the proposed hybrid framework encoder, both Residual and Swin Transformer blocks are combined for effectively capturing the multi-scale characteristics, such as irregular shapes and from small to large breast cancer images. Low-level spatial representations are first extracted by Residual stages, such as lesion contours, mass boundaries, micro-calcification clusters, and fine textural variations. Subsequently, these gathered locally enriched features from residual blocks are then forwarded to the Swin Transformer blocks, which model the long-range contextual relationships using W-MSA. Through this mechanism, the network can correlate distant but clinically relevant structures, for example, architectural distortions, asymmetries, or dispersed malignant tissue patterns, across the breast image. LN, an MLP, and residual connections are incorporated by each SwinT stage for stabilizing feature propagation and maintaining both local and global consistency. The MSCAS module refines intermediate feature maps by highlighting lesion region at multiple receptive fields prior passing them to the decoder. The decoder then reconstructs the spatial resolution through residual convolutional blocks and progressive up-sampling, ultimately generating a high-fidelity breast cancer segmentation mask of size $256 \times 256 \times 1$, accurately outlining malignant tissue regions.

### 3.1.3 Multi-Scale Channel Attention and Squeezing (Channel Aware)

The MSCAS block is introduced for capturing multi-scale channels from diverse receptive fields into a single discriminative descriptor. A multi-scale squeeze extracts both global and boundary-aware cues, which involves a lightweight excitation stage using 1×1 convolution, normalization, and nonlinear gating that leverages to emphasize channels tied to subtle tumor boundaries and heterogeneous tissue texture. SoftMax-normalized attention weights subsequently reweight the feature map, amplifying clinically meaningful channels while suppressing redundant or background-driven activations. Through this targeted recalibration, MSCAS improves hierarchical feature flow and substantially elevates the encoder's ability to represent malignant structures under dense-tissue and noise-prone imaging conditions. MSA module and an MLP, with both components preceded by LN and residual connections. This directed refinement will significantly increase the discriminative power of the encoder and is represented in Figure 3

$$X = X_b \oplus X_g \tag{8}$$

$$z_c = GAP(x)_c = \frac{1}{HW} \sum_{i=1}^{H} \sum_{j=1}^{W} x_c(i,j) \tag{9}$$

$$u = BN(w_{1x1} * x) \tag{10}$$

$$p = w_f z + b_f \tag{11}$$

$$s_c = \frac{e^{p_c}}{\sum_{j=1}^{C} e^{p_j}} \tag{12}$$

$$F_{ref} = s \odot Xs \tag{13}$$

This module initially combines complementary feature 'X' streams for capturing both sharp lesion borders and broadening the tissue structures, as elaborated in equation (8). The squeezed descriptor is generated for summarizing global information through channel-wise averaging GAP(.) as given in Equation (9). This descriptor is convolved by a 1×1 convolution stage, transformed using batch normalization, and a fully connected layer. Equations (10) and (11) mathematically represent these transformations. Finally, the SoftMax-based attention '$s_c$' weights reweigh the multi-scale features '$F_{ref}$' according to Equations (12) and (13).

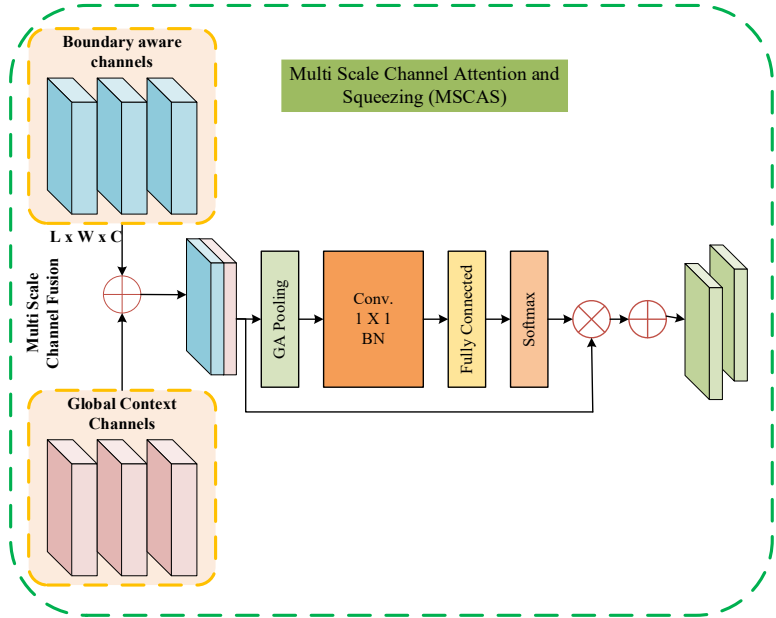

Figure 3: The proposed Channel Aware (MSCAS) block.

## 3.2 Multiscale Features Decoding

Target level and intermediate features achieved from the encoder side are provided to the decoder of the proposed 'Residual-SwinCA-Net' framework for reconstructing a high-resolution segmentation map. The decoding process began with the deep (16 × 16 × 512) feature map and feature fusion with the corresponding encoder stage, utilizing skip connections. The decoder consists of residual blocks, convolution blocks with an increasing number of filters, up-sampling 2D layers for feature maps refinement, and increasing spatial resolution progressively. Furthermore, MSCAS modules are incorporated at each stage for channel attention to the features coming from the encoder side to enhance the segmentation accuracy by allowing the framework to learn the relevant channels. Then, on corresponding stages, the channel feature maps are concatenated with the skip-connected, and the resolution is progressively refined and up-sampled sequentially (16 × 16 → 32 × 32 → 64 × 64 → 128 × 128) with channel dimension reduction. The final stage of the decoder consists of a convolution block of 40 filters, followed by up-sampling 2D for the final production of the (256 × 256 × 80) map, and will produce an output of (256 × 256 × 1) after further processing for the representation of the segmentation mask. This overall U-shaped encoder and decoder segmentation framework is shown diagrammatically in Figure 2.

### 3.2.1 Pixel Attention

This block will perform a pixel attention mechanism, which enhances the discriminative ability of the segmentation network by allowing it to focus on the most informative spatial locations within each feature map. Pixel attention helps in the discrimination of the lesion region by giving weight to the individual pixel, unlike channel attention, where emphasis was placed upon channel-level features. This kind of pixel attention is important due to the variant natures of the pixels with intra and inter class, irregular shapes, and low contrast boundaries, and is shown in Figure 4.

$$U_{sa\_out} = M_p Z_{enh} \quad (14)$$

$$h = \sigma_1(M_1 Z_{enh} + M_2 S_{m,n} + b_1) \quad (15)$$

$$M_p = \sigma_2(g(h) + b_2) \quad (16)$$

$$y = \sum_{i=1}^{A} \sum_{j=1}^{B} w_{i,j} u_{sa\_out} \quad (17)$$

$$\sigma_x = \frac{e^{x_i}}{\sum_{c=1}^{C} e^{x_c}} \quad (18)$$

The attention map finally passes through the PA module that produces the final $U_{sa\_out}$ through element-wise multiplication for amplification of the salient features space. Convolution, element-wise multiplication, and activation applied are shown in equations (17-18). The $U_{sa\_out}$ of the PA block will produce final and refined spatial features where the relevant region to breast cancer is enhanced. $S_{mn}$ showing the input features map is used as input, and the weighted features in the range of [0,1] are produced as $M_p$ in equation (14). Explanation of activation functions, biases, and transformations is given in equations (15 and 16), and the number of neurons and SoftMax activation are parameterized as equations (17 and 18).

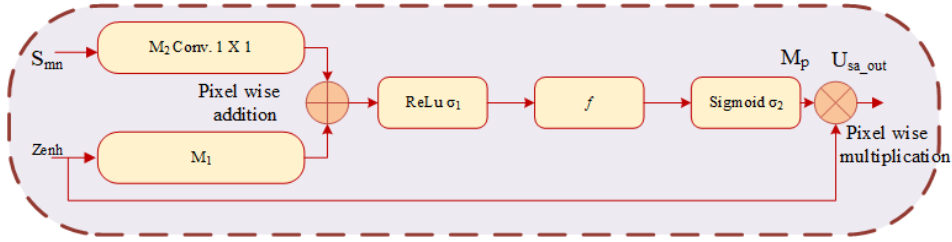

Figure 4: Spatial pixel Attention block.

Pixel attention will help the model in suppressing the background level information and highlighting the information related to tumor edges, micro-lesion clusters, and texture irregularities. An attention map is generated by giving high weights to the interclass pixel values and lowering the weights of intra-class pixels, and multiplying with the input feature maps, which will enhance pixel discrimination and noise reduction. This improved block will help the model in the discrimination of the similar-looking benign and malignant tissues, increase lesion level sensitivity, and improve boundary sharpness. Combining with the channel attention via MSCAS, this pixel attention will enhance both "where" (spatial focus) and "what" (channel focus). The information will be jointly optimized and will produce more accurate and better segmentation outcomes.

## 3.3 Implementation of Existing Segmentation CNN/ViTs

Establishing a strong benchmarking foundation, this work implemented a diverse set of state-of-the-art (SOTA) segmentation techniques, including pure CNNs, transformers such as ViTs, and hybrid CNN-

Transformers. The U-Shaped convolutional models, including U-Net++, UNeXt, MSRF-Net, ResUNet++, and Inf-Net++, are designed for capturing multi-scale contextual information through dense skip connections and hierarchical convolutions. These convolution U-Net networks work efficiently for local information but will struggle for learning long-range dependencies. Therefore, Transformer-based models such as Swin-UNet employ self-attention to capture wide contextual dependencies, offering stronger global reasoning at the cost of higher computation. Hybrid designs, including TransFuse and SwinBTS++, hybrid ViTs with CNNs to balance local detail and global structure. Our observations show that CNNs excel at fine-grained local discrimination, whereas transformers and hybrid variants benefit from their broader contextual modeling.

## 4 Experimental Setup

An 80:20 split defined the training–testing sets, and the training portion has been further partitioned into training–validation subsets. SOTA segmentation models have been selected for training and tested against the proposed model using the test data with 100 epochs. Optimizer, loss function, and learning rate were configured as SGD, cross-entropy, and $10^{-3}$. The dataset is passed through an augmentation process to make the model generalize.

### 4.1 Dataset

A dataset represented in Figure 5 is generated for this study by merging two publicly available datasets to test the robustness and improved performance of the proposed framework. The first one is BUSI [39] from Baheya Hospital, Egypt, comprising 780 images of 600 female patients, and the second is UDIAT from the UDIAT Diagnostic Center, Parc Taulí Corporation, Sabadell, Spain, which consists of clinically validated scans. All the images were passed through a preprocessing phase for compatibility with the model input dimension, such as 256× 256 × 3.

### 4.2 Implementation Details

All training experiments were executed on an Ubuntu 16.04 system powered by an Intel Xeon CPU and an NVIDIA GeForce GTX 1080Ti GPU, implemented within the PyTorch framework [42]. The Adam optimizer was utilized with a base learning rate of $1e^{-4}$ and default momentum terms, batch size of 16 for 100 epochs, while the learning rate schedule is controlled using LambdaLR. To improve generalization, all images had (256 × 256) pixels and passed through augmentation methods such as flipping and rotation to simulate the variability of tumor appearances in clinical scenarios. To enhance generalization and prevent overfitting due to limited data, we applied on-the-fly augmentations during each training cycle, including center cropping, horizontal and vertical flips, and random rotations of up to 10°.

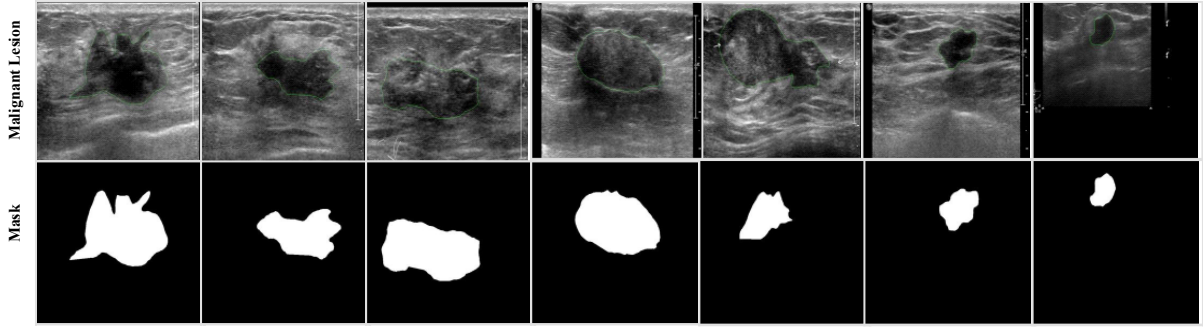

Figure 5: BUSI dataset showing a malignant tumor with a mask.

## 4.3 Loss Function and Performance Metrics

In segmentation tasks, binary cross-entropy (BCE) loss is utilized. To tackle class imbalance in segmentation between the foreground (object) and background, a Dice coefficient-based loss is applied. This metric, widely regarded as a reliable measure of dataset similarity, serves as an effective indicator of segmentation accuracy. The Dice segmentation loss $L_{dice}$ quantifies the similarity between the ground truth (GT) and $Y_{seg}$ the predicted segmentation map $P_{seg}$. To enhance segmentation results, LossNet was integrated into the polyp segmentation framework from [44], enabling precise control over features and structures at the feature level. The segmentation loss is defined as follows:

$$L_{dice}(P_{seg}, Y_{seg}) = 1 - \frac{2P_{seg}Y_{seg}+1}{P_{seg}+Y_{seg}+1} \tag{19}$$

$$L_{BCE} = -\frac{1}{N}\sum_{i=1}^{N}[y_i \log(p_i) + (1-y_i)\log(1-p_i)] \tag{20}$$

$$L_{Dice} = 1 - \frac{2\sum_{i-1}^{N} y_i p_i}{\sum_{i-1}^{N}(y_i+p_i)} \tag{21}$$

$$L = \frac{L_{BCE}+L_{Dice}}{2} \tag{22}$$

The ground-truth label indicates whether each $i^{th}$ pixel is a member of the positive class, whereas $N$ denotes pixels in the input image $y_i \in \{0,1\}$. The predicted probability $p_i \in \{0,1\}$ The model's confidence indicates how confident it is in classifying pixels. Utilizing the stability of Cross-Entropy and the efficacy of Dice loss in managing imbalance, a hybrid loss function combining the two is used to solve class imbalance and enhance segmentation accuracy. Segmentation performance was assessed across all breast ultrasound datasets using standard metrics: accuracy, intersection over union (IoU), and Dice similarity coefficient (DSC), defined in equations 23-25: In these formulations, TP and TN denote the counts of true positive and true negative pixels, respectively, while FP and FN refer to false positives and false negatives.

$$DSC = \frac{2 \times TP}{2 \times TP + FP + FN} \tag{23}$$

$$Acc = \frac{TP+TN}{TP+TN+FP+FN} \tag{24}$$

$$IoU = \frac{TP}{TP+FP+FN} \qquad (25)$$

## 5 Results and Discussions

The proposed Residual-SwinCA-Net achieved optimal performance across lesion and background regions. BUSI Lesion DSC improves from 0.8894 to 0.9041 and IoU from 0.9521 to 0.96735, indicating more reliable malignant region analysis as shown in Table 1. These region-wise improvements align with the global metrics, where GlobalAcc increases to 0.9933, MeanIoU reaches 0.9771, and MeanBFScore improves to 0.9793, as illustrated in Table 2. Compared with hybrid CNN-ViTs, such as SwinBTS++ (DSC 0.8611, IoU 0.91835), and standalone ViTs/CNNs, including Swin-UNet (DSC 0.8015) and ResUNet++ (DSC 0.7895), the integrated Residual-SwinCA-Net achieves substantially higher lesion fidelity. The framework integrates a customized residual CNN for preserving fine-grained local correlations with a customized Swin encoder that reinforces long-range contextual and structural discrimination. The MSCAS block further enhances multi-scale feature coherence through adaptive channel attention, suppressing irrelevant responses and stabilizing boundary continuity. Consequently, the Residual-SwinCA-Net achieves significant improvements in DSC (4.3–22.5%), IoU (4.9–31.72%), and BFScore (9.62–43.93%), indicating more accurate region overlap and superior global-local feature integration.

Table 1. Quantitative comparison of the proposed framework with existing CNNs/ViTs.

| Techniques | GlobalAcc. | MeanAcc. | Mean IoU | WeightedIoU | MeanBFScore |
|---|---|---|---|---|---|
| **Proposed Residual-SwinCA-Net** | 0.9933 | 0.9929 | 0.9771 | 0.98454 | 0.9793 |
| **Proposed Residual-Swin-Net** | 0.9910 | 0.9903 | 0.9712 | 0.9793 | 0.9621 |
| **Existing Hybrid CNN-ViT** | | | | | |
| SwinBTS++ | 0.9895 | 0.9883 | 0.9532 | 0.9798 | 0.9104 |
| TransFuse | 0.9606 | 0.9488 | 0.84745 | 0.9298 | 0.6861 |
| UNeXt | 0.9557 | 0.9327 | 0.8297 | 0.9215 | 0.6785 |
| **Existing ViTs/CNNs** | | | | | |
| Swin-Unet | 0.9768 | 0.9488 | 0.8993 | 0.9561 | 0.7936 |
| Inf-Net++ | 0.9760 | 0.9481 | 0.8982 | 0.9555 | 0.7929 |
| ResUNet++ | 0.9737 | 0.9819 | 0.8944 | 0.9520 | 0.7799 |
| MSRF-Net | 0.9631 | 0.9371 | 0.8519 | 0.9331 | 0.7070 |
| DeepLabV3 | 0.9548 | 0.9738 | 0.8372 | 0.9220 | 0.6980 |
| U-Net++ | 0.9366 | 0.9577 | 0.7891 | 0.8949 | 0.5997 |

The proposed Residual-SwinCA-Net has shown clear improvement by incorporating MAS DSC lesion from 0.8894 to 0.9041 (1.65% gain), while the IoU rises from 0.9521 to 0.96735 (1.60%). Moreover, the performance gain of the proposed Residual-SwinCA-Net over the existing hybrid CNNs and ViT is illustrated in Figure 6. Finally, a detailed qualitative and visual segmentation comparison of the proposed framework and the SOTA models is presented in Figures 7 and 8. These figures show the original input and the ground truth mask in the first two columns, and the segmented areas are represented by the color and binary regions. The similarity among these colored predicted regions and ground-truth masks demonstrates the segmentation accuracy. Predicted masks generated by the proposed framework can be clearly seen near resembling those of the original masks. The effectiveness of the proposed framework can be observed by accurately delineating the lesion boundaries and preserving fine structural details, even with the irregular shape and complex regions.

Table 2. Quantitative comparison in terms of segmentation metrics.

| Techniques | Regions | DSC | Accuracy | IoU | BFScore |
|---|---|---|---|---|---|
| **Proposed Residual-SwinCA-Net** | Lesion | 0.9041 | 0.99111 | 0.96735 | 0.97668 |
|  | background | 0.9887 | 0.99467 | 0.99087 | 0.98193 |
| **Proposed Residual-Swin-Net** | Lesion | 0.8894 | 0.99920 | 0.9521 | 0.95264 |
|  | background | 0.9827 | 0.99121 | 0.9896 | 0.95233 |
| **Existing Hybrid CNN-ViT** | | | | | |
| SwinBTS++ | Lesion | 0.8611 | 0.98671 | 0.91835 | 0.88049 |
|  | background | 0.9871 | 0.98989 | 0.98811 | 0.94033 |
| TransFuse | Lesion | 0.7358 | 0.93326 | 0.73921 | 0.60797 |
|  | background | 0.969 | 0.96434 | 0.95568 | 0.76049 |
| UNeXt | Lesion | 0.7267 | 0.90237 | 0.70896 | 0.59439 |
|  | background | 0.9659 | 0.96294 | 0.95034 | 0.7588 |
| **Existing ViTs/CNNs** | | | | | |
| Swin-Unet | Lesion | 0.8015 | 0.91205 | 0.82465 | 0.71937 |
|  | background | 0.9780 | 0.9856 | 0.97397 | 0.86534 |
| Inf-Net++ | Lesion | 0.6943 | 0.83205 | 0.69465 | 0.71937 |
|  | background | 0.9611 | 0.9856 | 0.94397 | 0.86534 |
| ResUNet++ | Lesion | 0.7895 | 0.9026 | 0.81857 | 0.73733 |
|  | background | 0.9807 | 0.97112 | 0.97015 | 0.82241 |
| MSRF-Net | Lesion | 0.7508 | 0.88302 | 0.74524 | 0.62671 |
|  | background | 0.9701 | 0.97124 | 0.95862 | 0.78372 |
| DeepLabV3 | Lesion | 0.7203 | 0.85863 | 0.72562 | 0.67852 |
|  | background | 0.9691 | 0.94889 | 0.94872 | 0.71746 |
| U-Net++ | Lesion | 0.6791 | 0.81552 | 0.65017 | 0.53738 |
|  | background | 0.9565 | 0.92994 | 0.92811 | 0.66201 |

## 5.1 Comparative Analysis with Hybrid Transformers

The proposed Residual-SwinCA-Net segmentation framework is compared with other SOTA techniques, demonstrating good performance in segmenting the BUSI lesion showing in Figure 7. Table 2 shows that the performance differences between the selected hybrid and transformer techniques have been achieved due to residual CNN blocks that capture low-level spatial and boundary details effectively. The long-range contextual dependencies are modeled through SwinT blocks along with the MSCAS block. Potential segmentation gaps are mitigated due to this attention mechanism, which slightly reduces the performance of pure Transformer/hybrids during feature concatenation. Compared with the other SOTA models, such as SwinBTS++ and TransFuse, the proposed model achieves consistent performance. Due to the lack of an effective cross-scale features recalibration mechanism in SwinBTS++ will be limited to capturing boundary-level and contextual representation. On the other hand, TransFuse incorporates a dual-branch design using convolution and transformer streams, but lacks systematic integration and a unified attention mechanism for aligning features across scales will result in inferior performance. Its independent fusion strategy often leads to spatial inconsistency and reduced sensitivity to subtle texture variations present in medical images.

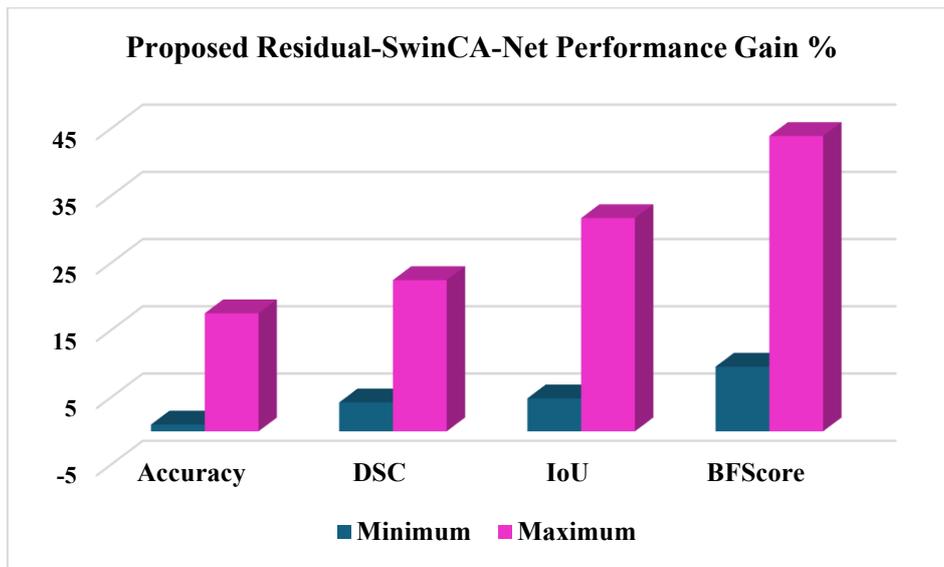

Figure 6: The proposed framework's comparison with the SOTA segmentation ViTs/CNNs.

## 5.2  Comparison with CNN and Multi-Scale Techniques

ResUNet++ outperforms the existing CNN, which is the modified version of the traditional UNet, incorporating residual and squeeze-and-excitation blocks. ResUNet++ focuses primarily on local spatial features but lacks global context, which will lead to misclassification in regions with weak boundaries. Likewise, MSRF-Net is designed with convolutional receptive fields for multiscale feature fusion that struggle to capture long-range information for learning complex medical images and texture variation. Conversely, Lightweight token-mixing blocks replace the standard convolution operations by UNeXt, making it computationally efficient, but compromising fine-grained feature extraction and having insufficient capacity to handle the subtle intensity variations and irregular shapes. In this regard, the proposed Residual-SwinCA-Net learn variation, global context, and morphology by systematically integrating global and local fine-grained feature capturing blocks. Therefore, the proposed framework achieved performance gain, DSC (4.3-22.5%), Accuracy (8.85-22.5%), IoU (14.88-31.72%), and BF Score (23.94-43.93%) over CNN and multi-scale techniques in the BUSI lesion area segmentation (Tables 1 & 2) and Figures 6-8. Comparison of the proposed and previous work is given in Table 3.

Table 3. Comparison with the previous methods employed on the BUSI dataset.

| Techniques | DSC | Accuracy | IoU |
|---|---|---|---|
| **Proposed Residual-SwinCA-Net** | 0.9041 | 0.99111 | 0.9674 |
| **Proposed Residual-Swin-Net** | 0.8894 | 0.99920 | 0.9521 |
| Existing Hybrid CNN-ViT | | | |
| TransUNet [43] | 0.7753 | 0.9642 | --- |
| UCTransNet [44] | 0.7828 | 0.9631 | --- |
| LCA-Net [45] | 0.6721 | 0.9323 | -- |
| MultiResNet [46] | 0.7798 | 0.9667 | --- |
| UDTransNet [47] | 0.8019 | 0.9677 | 0.6813 |
| Existing CNNs | | | |
| FCN [48] | 0.7676 | 0.9565 | --- |
| MSDANet [49] | 0.6845 | 0.9566 | --- |
| DAU-Net [50] | 0.7423 | 0.9621 | --- |
| Attention UNet [51] | 0.7583 | 0.9654 | --- |
| AAU-Net [52] | 0.8023 | 0.9669 | 0.6912 |
| BMFNet [53] | 0.8347 | 0.9681 | 0.7113 |

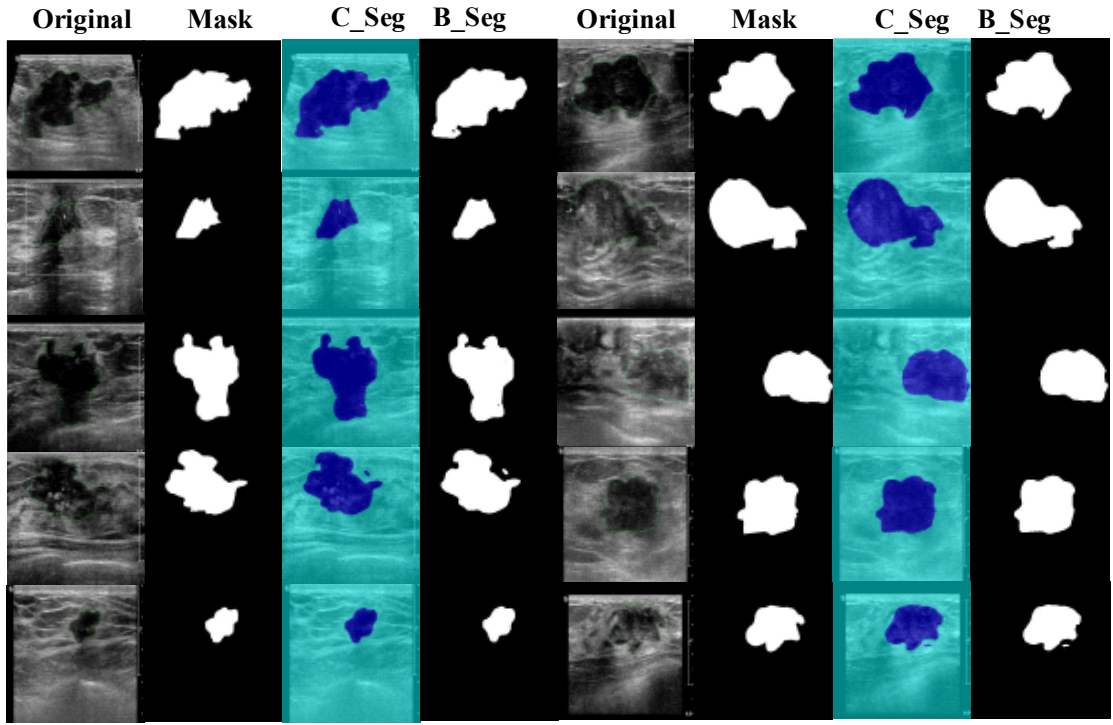

Figure 7: Segmented BUSI lesion outcome (color and binary) of the proposed Residual-SwinCA-Net.

## 5.3 Segmentation Quality and Localization

The most critical measures for assessing the spatial overlap between the predicted mask and the ground truth, as well as evaluating the quality of segmentation, can be calculated through IoU, Mean IoU, and Weighted IoU (Figure 8). These outcomes are given in Table and Figures 8-9, showing that the proposed framework is very effective in terms of localization and boundary precision due to a hybrid feature extraction and the contribution of the MSCAS block on various levels. Due to limited receptive fields,

CNNs such as U-Net++, MSRF-Net, and ResUNet++ exhibit lower IoU scores. Also, these techniques have lower capabilities to model long-range dependencies will result in fragmented and incomplete segmentation in the complex regions. Standard IoU is sensitive to both false positives (predicting background as foreground) and false negatives (missing actual lesion pixels) in the confusion matrix (Table 4), showing significant improvement of the proposed framework over SOTA techniques. Specifically, on the merged dataset, the proposed framework achieves accurate pixel classification of both lesion and background, demonstrating its effectiveness in the confusion matrix Table 4.

Table 1, 2, and Figure 9 present the DSC and Mean Boundary F-score (MeanBF) for segmenting the target lesion and accurately delineating its boundaries, high MeanBF indicating effective boundary localization. The proximity of the predicted boundary evaluated by MeanBF analysis to the true boundary, yielding results that indicate perfect boundary localization. The incorporation of effective residual connections throughout the U-shaped structure and systematic implementation of LoG operation ensures robust gradient flow, preserving crucial high-resolution boundaries. Moreover, the MSCAS module ensures that the decoder receives features that are optimally prepared for precise boundary prediction. These background and lesion improvements can be seen graphically in Figure 7, 9, and 10, and the confusion matrix in Table 4.

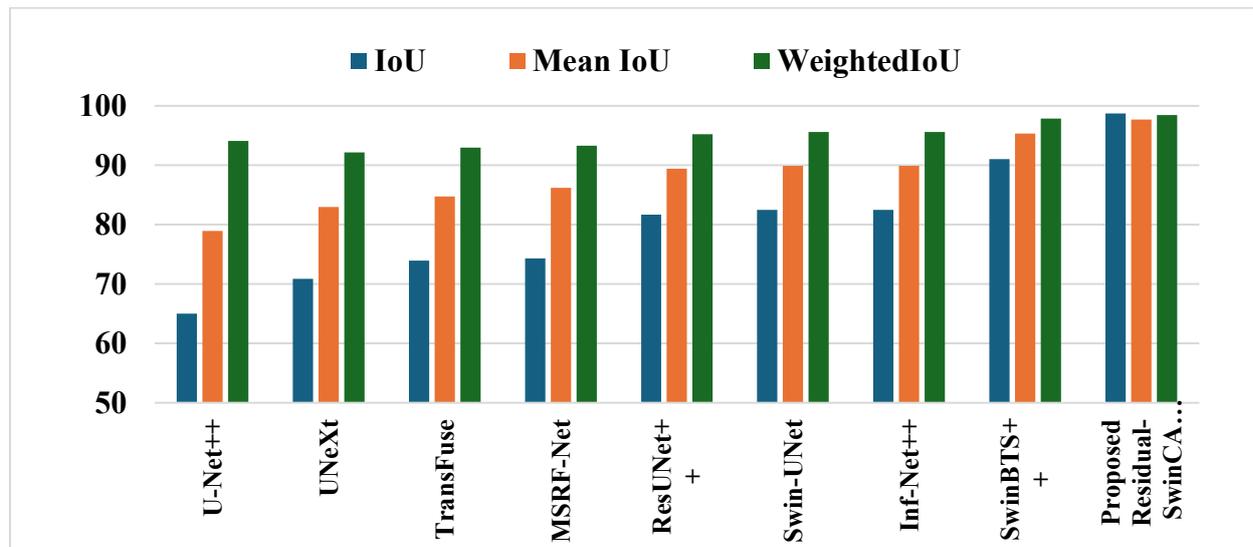

Figure 8: The proposed and existing Segmentation CNNs/ViTs on spatial overlap metrics.

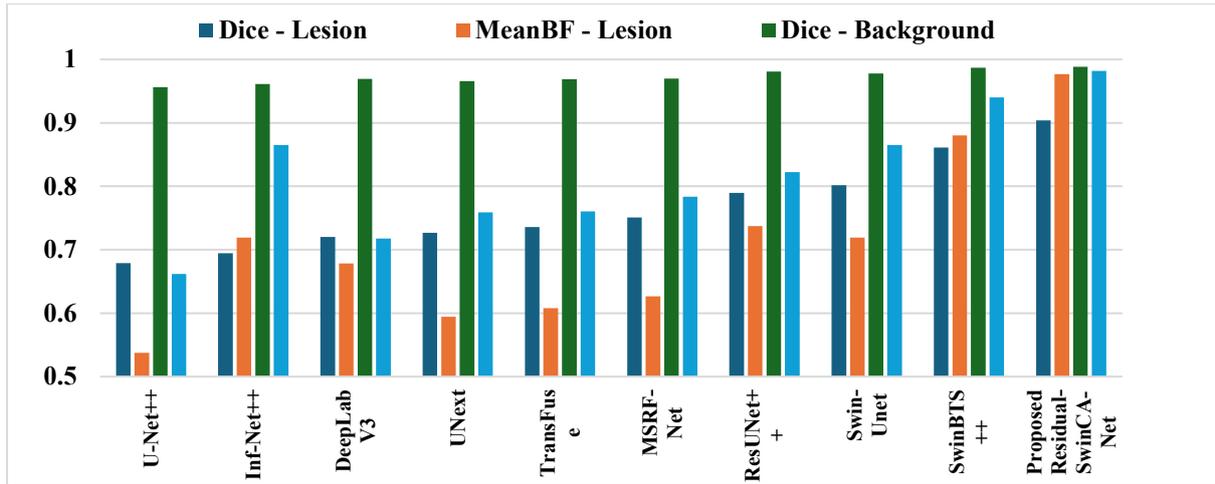

Figure 9: Comparative analysis of the proposed and existing techniques using the DSC and MeanBF metrics for both the lesion and background categories.

Table 4. Comparison of the proposed framework in terms of pixel classification.

| Models | Confusion Matrix | Lesion | background |
|---|---|---|---|
| **Proposed Residual-SwinCA-Net** | Lesion | 0.98017 | 0.0158258 |
| | background | 0.0011572 | 0.99184 |
| **Proposed Residual-Swin-Net** | Lesion | 0.97851 | 0.02149 |
| | background | 0.00990 | 0.99010 |
| Existing Hybrid CNN-ViT | | | |
| SwinBTS++ | Lesion | 0.97671 | 0.023295 |
| | background | 0.01011 | 0.98989 |
| TransFuse | Lesion | 0.93326 | 0.06674 |
| | background | 0.035656 | 0.96434 |
| UNeXt | Lesion | 0.90237 | 0.097627 |
| | background | 0.037058 | 0.96294 |
| Existing ViTs/CNNs | | | |
| Swin-Unet | Lesion | 0.96205 | 0.037951 |
| | background | 0.014396 | 0.9856 |
| Inf-Net++ | Lesion | 0.95262 | 0.0473837 |
| | background | 0.028881 | 0.97112 |
| ResUNet++ | Lesion | 0.94262 | 0.0573837 |
| | background | 0.028881 | 0.97112 |
| MSRF-Net | Lesion | 0.90302 | 0.096985 |
| | background | 0.028756 | 0.97124 |
| DeepLabV3 | Lesion | 0.90863 | 0.0913739 |
| | background | 0.051105 | 0.94889 |
| U-Net++ | Lesion | 0.89552 | 0.104484 |
| | background | 0.07006 | 0.92994 |

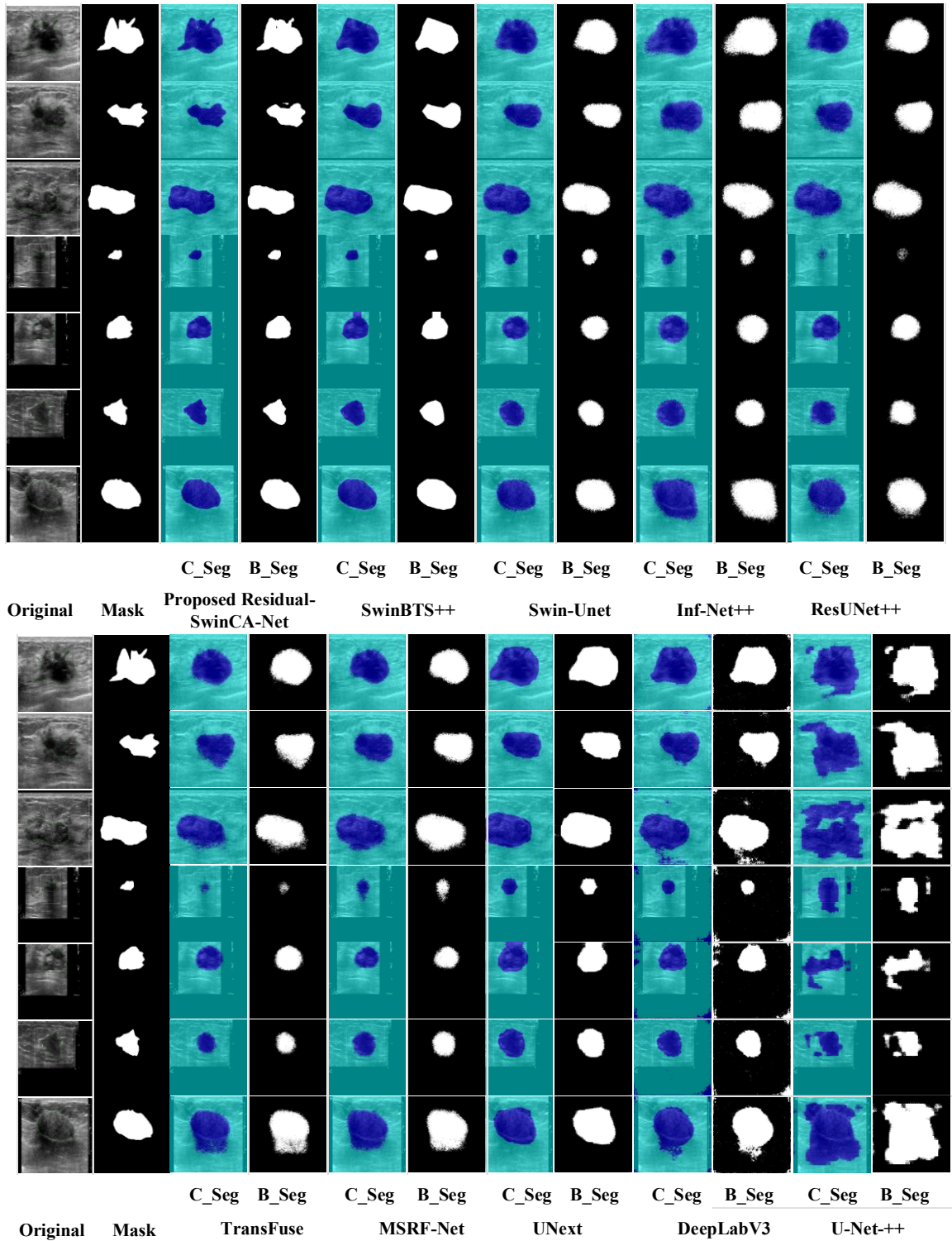

Figure 10: BUSI lesion segmentation analysis of the proposed framework with SOTA CNNs/ViTs.

## 6 Conclusion

This study incorporated an integrated residual CNN and customized SwinT in a unified segmentation network that extracts both local and global information from the BUSI dataset. The dataset included speckle noises removed by LoG application used after residual CNN and SwinT blocks, resulting the enhanced contrast, sharpened edges, emphasizing the important structures, and providing discrimination between lesion and background regions. Additionally, at every decoding stage in Residual-SwinCA-Net, MSCAS block, is incorporated for adaptively integrating information at multiple receptive fields and focuses on critical channel-wise responses of features. Furthermore, the discriminative power of segmentation is boosted by the pixel attention mechanism by enabling the Residual-SwinCA-Net to concentrate on the most spatially informative locations within each feature map. The model is evaluated for performance measure and applicability by the generated dataset, which shows a Dice score of 0.9041, an IoU of 98.74%, and a MeanBF of 0.9819. These outcomes have shown the effectiveness of the Residual-SwinCA-Net framework in improving localization and boundary accuracy with balanced local details and global contextual information. Overall, the proposed framework has the potential for integration into computer-based diagnosis systems, in early detection of the breast cancer regions for clinicians with the highest confidence.

**Future directions**

A dynamic fusion strategy and lightweight attention mechanisms are required to reduce computational costs and usage in real-time clinical systems. Further, semi and self-supervised learning approaches will overcome data scarcity and efficient learning in imbalanced medical datasets while enhancing robustness. MRI and ultrasound images will be used for further improvement of accuracy and diagnosis to effectively improve the adaptability of the system in the real world.